%% file: main.tex
\ifdefined\XeTeXversion\else\pdfoutput=1\fi
\documentclass{article}

\usepackage{arxiv}

\usepackage[utf8]{inputenc}
\usepackage[T1]{fontenc}
\usepackage{hyperref}
\usepackage{url}
\usepackage{booktabs}
\usepackage{array}
\usepackage{multirow}
\usepackage{amsmath,amssymb,amsfonts}
\usepackage{nicefrac}
\usepackage{microtype}
\usepackage{graphicx}
\usepackage{xcolor}
\usepackage{natbib}
\usepackage[capitalise]{cleveref}
\crefname{figure}{Fig.}{Figs.}
\Crefname{figure}{Figure}{Figures}
\usepackage{framed}
\usepackage{algorithm}
\usepackage{algpseudocode}
\usepackage{caption}
\usepackage{placeins}
\hypersetup{colorlinks=true, linkcolor=[HTML]{1F4E8C}, citecolor=[HTML]{1F4E8C}, urlcolor=[HTML]{1F4E8C}}
\newcommand{\Phiz}{\Phi_0}
\newcommand{\coderepo}{\url{https://github.com/Fibonaccirabbit/GroundedRSI}}
\title{Grounded Recursive Self-Improvement\\ of Robot Agents}

\author{
  Zhanjiang Yang \\
  Shenzhen University \\
  Shenzhen, China \\
  \texttt{zhanjiangyang3@gmail.com}
  \And
  Xinan Xu \\
  University of Science and Technology of China \\
  Hefei, China \\
  \texttt{xuxinan@mail.ustc.edu.cn}
}

\renewcommand{\shorttitle}{Grounded Recursive Self-Improvement of Robot Agents}

\date{}

\begin{document}
\maketitle

\begin{center}
  \includegraphics[width=\linewidth]{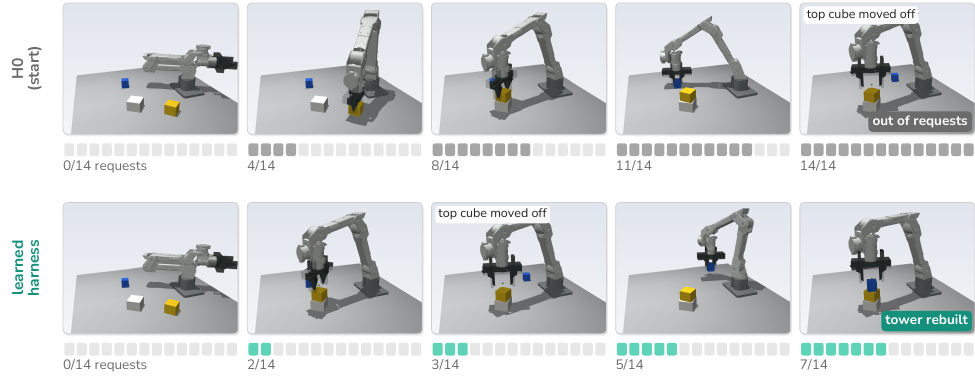}
  \captionof{figure}{The same held-out task under the starting harness H0 (top) and under a harness the agent learned for itself (bottom). Bars: model requests used, of a budget of 14; the learned harness finishes where H0 runs out.}
  \label{fig:teaser}
\end{center}
\vspace{6pt}

\begin{abstract}
Robots increasingly carry out tasks on their own, yet making them better is still largely driven by people. Recursive self-improvement would let a robot improve both how it acts and how it improves, without giving up safety. We ask how far a robot agent can go toward this when the only thing it may change is its own harness, the text its frozen language model reads. We grade the answer on a recent six-level scale of physical recursive self-improvement. Our agent drives a simulated six-axis arm through nine low-level tools. A proposer rewrites the rules and facts in its harness, each fact grounded in a measurement the robot made; a static gate and a paired validation decide which rewrites are kept, and every gain is measured on task families the procedure never sees. The answer: improvement that carries over to unseen task families, yes (L3); improving the improvement procedure, one promoted step whose advantage is not resolved over runs (L4); verified and safe recursive gain, not shown (L5). Kept harnesses pass 18 to 42 more of 99 held-out episodes than the starting harness, and under a fixed request budget what they learn is mainly request economy, not body knowledge. At the meta level the procedure first drifted toward the body; a rewrite was promoted only after we added a gate that keeps the method robot-agnostic. Every held-out table contact came from an improved harness and from one finger part that no tool reports, so such hazards must be protected constraints. Code is available at \coderepo.
\end{abstract}

\input{sections/intro}

\input{sections/related}

\input{sections/definitions}

\input{sections/system}

\input{sections/method}

\input{sections/protocol}

\input{sections/l3}

\input{sections/l4}

\input{sections/l5}

\FloatBarrier
\input{sections/discussion}

\input{sections/conclusion}

\bibliographystyle{unsrtnat}
\bibliography{references}

\input{sections/appendix}

\end{document}

%% file: sections/intro.tex
\section{Introduction}
\label{sec:intro}

Robots increasingly carry out tasks on their own, but making them better is still largely driven by people: people collect and review the data, diagnose failures and decide which update to adopt \citep{zhou2026whoimproves}.
A robot that improved itself would turn its own experience into lasting updates.
One that improved itself \emph{recursively} would also improve the procedure that makes those updates.
\citet{zhou2026whoimproves} make these notions precise for physical systems and grade them in six levels, from L0 (no autonomous retained change) to L5 (recursive gain verified by an independent evaluator).
How far a concrete robot system can climb these levels is an open question.
We ask it for the simplest system we could build: a robot agent built on a frozen large language model (LLM) that may change only its own \emph{harness}, the text the model reads.

The question splits into three, one per level we test.
Does the agent improve itself across tasks (L3)?
Does it improve how it improves (L4)?
Is that recursive gain verified and safe (L5)?
Each needs its own evidence.
The procedure that makes changes chooses them on the tasks it practises, so a gain there may be noise, may not carry over to other tasks, or may be bought with unsafe actions.
A better procedure, in turn, must be shown to produce better systems from the same starting point and budget.
On a robot, trials are costly and some mistakes cannot be undone.

\begin{figure}[t]
  \centering
  \includegraphics[width=\linewidth]{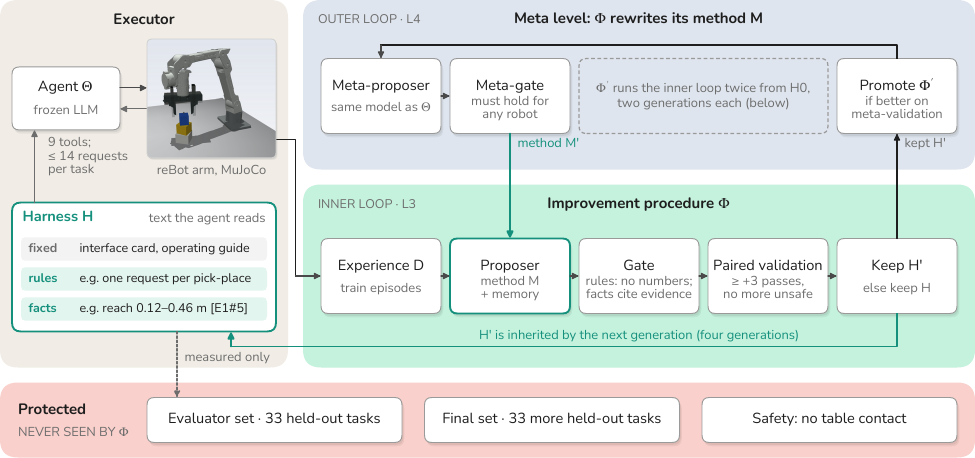}
  \caption{The system: a frozen agent reads its harness (left); the improvement procedure changes the harness (middle, L3) and, at the meta level, its own method (top, L4); the held-out sets and the safety condition are protected (bottom).}
  \label{fig:system}
\end{figure}

Our approach keeps every retained change small enough to audit and every claim cheap enough to check.
The only thing that changes is text: \emph{rules} about how to act, which may contain no numbers, and \emph{facts} about the body, each of which must cite the tool result that measured it.
A proposer writes candidate rules and facts from the agent's own episodes; a static gate enforces these constraints; and a paired validation keeps (\emph{promotes}) a candidate only if it passes more episodes without being less safe (\cref{fig:system}).
The proposer follows a \emph{method text}, which the system may also rewrite, and that is a change to the improvement procedure itself.
Every gain is measured on task families held out by skill sequence, in two sets the procedure never sees.
We call self-improvement of this kind \emph{grounded}, in the sense of the definitions' requirement that retained changes be physically grounded: every fact the harness keeps cites what the robot measured (the grounding split of \cref{sec:method}), and every change is kept only after it has been executed and validated in the physics simulator.

For one robot agent in simulation, the answer to this question is: across tasks, yes (L3); improving how it improves, one promoted step whose advantage is not resolved over runs (L4); verified and safe recursive gain, not shown (L5).
For L3, the harnesses kept by five improvement runs pass 18 to 42 more of 99 held-out episodes than the starting harness H0.
\Cref{fig:teaser} shows one such episode: the simulator moves the top cube off the tower after the second placement, and the harness learned by the first run of the procedure's third version completes the task, rebuilding the tower, within six model requests and answers in the seventh, where H0 runs out.
For L4, after a gate we added required the method text to name nothing specific to the robot, one rewrite was promoted; in fresh runs its harnesses scored higher on the held-out sets, but with three runs per method the advantage is not resolved.
For L5, those harnesses touched the table in two held-out episodes and those of the original method in none, which fails a safety criterion we registered in advance.

Three findings explain where the climb stops.
\begin{itemize}
  \item \textbf{Request economy, not embodiment.} Under a fixed budget of model requests, the kept harnesses make at least as many robot calls (tool calls) as H0 but need 14 to 26\% fewer requests, while a human-written description of the body (H$^*$) helps little. The economy is in deliberation, not in time, and it did not carry over to the one other executor model we tried (\cref{sec:l3}).
  \item \textbf{Left alone, the meta level drifted toward the body.} In the first of two meta-level attempts, the procedure filled its method text with operating instructions for this robot; only a gate we wrote, requiring the method to hold for any robot, kept the change at the level of learning (\cref{sec:l4}).
  \item \textbf{Safety facts the tools cannot see must be protected.} All seven table contacts on the held-out sets came from improved harnesses and from the same finger part, which sits 12\,mm below the tip that the tools report; six came at a commanded tip height of 12\,mm. The agent cannot observe this, so the motion layer, not the harness, has to enforce the clearance (\cref{sec:l5}).
\end{itemize}

In summary, we contribute
(i) a measured answer to how far a harness-only robot agent climbs the levels of physical recursive self-improvement (\cref{sec:l3,sec:l4,sec:l5});
(ii) evidence that, under a fixed request budget, what such an agent learns is request economy, which ties the harness to its executor;
(iii) the two lessons named above for the meta level and for safety;
and (iv) a protocol that makes each level claim checkable: splits by skill sequence, a hidden evaluator, measured noise, conclusive-episode accounting, and pre-registered criteria for the meta level (\cref{sec:protocol}).

%% file: sections/related.tex
\section{Related work}
\label{sec:related}

\paragraph{Recursive self-improvement.}
The idea that a machine could design better machines, including better versions of itself, goes back to \citet{good1966speculations}; the G\"odel machine of \citet{schmidhuber2007goedel} makes it formal, rewriting any part of its own code once it can prove that the rewrite is useful.
With large language models the idea has become something one can run.
STOP improves a program that improves programs by applying that program to itself \citep{zelikman2024stop}; Promptbreeder evolves both task prompts and the prompts that mutate them \citep{fernando2024promptbreeder}; the G\"odel Agent lets an agent modify its own logic \citep{yin2025godelagent}; and automated design of agentic systems \citep{hu2025adas} and the Darwin G\"odel Machine \citep{zhang2026dgm} search over agent designs, the latter keeping an archive of agents that each rewrite their own code.
\citet{zhou2026whoimproves} carry the notion over to physical systems, where trials are costly and some errors cannot be undone, and grade it in six levels.
Our meta level follows the same pattern as these systems, in that the method text that writes the harness is itself rewritten; what we add is a measurement of one such system on a robot against the levels.

\paragraph{Robots that improve themselves.}
Robots have long been asked to keep learning after deployment \citep{thrun1995lifelong,lesort2020continual,liu2023libero}.
One line of work lets them practise and update their own policies: by autonomous reinforcement learning with resets learned alongside the task \citep{sharma2023selfimproving}, by fine-tuning navigation policies on experience gathered in the wild \citep{hirose2024selfi,stachowicz2024lifelong}, by training on data that the robot's own policy generated \citep{bousmalis2023robocat}, by collecting and relabelling practice with foundation models \citep{zhou2024autonomous}, by self-improvement of embodied foundation models after imitation \citep{ghasemipour2025selfimproving}, and by improving a policy inside a learned world model \citep{yang2026compositional}.
A second line uses language models to write what the robot learns from: reward functions \citep{ma2024eureka}, simulated tasks and scenes \citep{wang2024robogen}, training environments and curricula \citep{liang2024curriculum}, and libraries of reusable skills \citep{wang2024voyager,tziafas2024library}.
In our study the language model stays frozen and nothing is trained; the only retained change is the text of the harness, which keeps every change readable and lets us check it before it is kept.

\paragraph{Language-model agents that act on robots.}
Large language models have been used to choose robot skills by their affordances \citep{ahn2022saycan}, to replan from feedback \citep{huang2022innermonologue}, and to write robot programs that call perception and control functions \citep{liang2023codeaspolicies,singh2023progprompt}.
Agents that interleave reasoning with tool calls \citep{yao2023react} can also issue several calls in one step, as code \citep{wang2024codeact} or as a parallel plan \citep{kim2024llmcompiler}.
Our agent acts in this way through low-level tools, and what its harness learns turns out to be largely how to pack robot calls into fewer model requests.

\paragraph{Text as the retained change.}
Language agents can improve from their own experience by writing it down: reflections that are carried into the next attempt \citep{shinn2023reflexion}, insights extracted from many tasks \citep{zhao2024expel}, or instructions optimised from scored attempts \citep{yang2024opro}.
Our harness is of this kind, with two constraints taken from the physical setting: rules may contain no numbers, and every fact about the body must cite the tool result that measured it.

%% file: sections/definitions.tex
\section{Definitions and what counts as evidence}
\label{sec:definitions}

We follow the definitions of \citet{zhou2026whoimproves} throughout and restate the parts our claims rest on.
Where their text leaves a choice open, we say which choice we made and why.

\subsection{The improving system}

The state retained after $k$ rounds of execution and improvement is the $k$-th generation
\begin{equation}
  S_k = (\Theta_k, H_k, \Phi_k, D_k), \qquad \Phi_k = (\mathrm{Adapt}_k, \mathrm{Train}_k, \mathrm{Ver}_k),
\end{equation}
where the model $\Theta_k$ and the harness $H_k$ form the task-execution system, the three stages of $\Phi_k$ form the improvement procedure, and $D_k$ is the store of experience and feedback.
In each round the current generation executes tasks, its trajectories enter $D$, and $\Phi_k$ maps the generation to the next one:
$(\Theta_{k+1}, H_{k+1}, \Phi_{k+1}) = \Phi_k(\Theta_k, H_k, \Phi_k; D_{k+1})$.
Because $\Phi_k$ is one of its own arguments, a change may reach the improvement procedure itself.

A component belongs to $H$ or $\Theta$ if the task-execution system uses it while carrying out a task, and to $\Phi$ if it is used only to produce or judge retained changes.
A change is \emph{inherited} when $\Phi$ keeps it on evidence in $D$ and later generations start from it.

The \emph{independent evaluator} $E$ is a set of tasks with a scoring criterion $Q$ that $\Phi$ cannot read, train on, or modify.
It scores the task-execution system, written $J_Q(\Theta_k, H_k)$; the improvement procedure is evaluated only indirectly, through the systems it produces.
In our experiments $E$ consists of two held-out sets, the \emph{evaluator set} (from here on, the evaluator) and the \emph{final set} (\cref{sec:protocol}).
$E$ and the safety rules together form the \emph{protected constraints}, which are kept out of reach of $\Phi$.

\subsection{Self-improvement and recursive self-improvement}

\paragraph{Definition 1 (physical self-improvement).}
A system improves itself over rounds $1,\dots,K$ if
(i) at least one change to $\Theta$ or $H$ is produced by stages of $\Phi$ and inherited, and no human writes into $\Theta$, $H$, or $\Phi$;
(ii) the inherited changes on which the retained gain depends are physically grounded, ``whether in a physics simulator or in the real world'';
and (iii) the gain is retained: $J_Q(\Theta_K, H_K) > J_Q(\Theta_0, H_0)$.

\paragraph{Definition 2 (physical recursive self-improvement).}
Self-improvement is recursive if, in addition,
(i) $\Phi$ is able to update its own stages, so that changes to $\Phi$ produced by $\Phi$ are inherited and used by later rounds;
(ii) these changes are physically grounded as above;
and (iii) starting from the same generation and experience, and under the same budget, the system produced by each updated $\Phi_k$ scores a higher $J_Q$ than the system produced by $\Phi_{k-1}$, with the gain measured across tasks.

\paragraph{Tasks.}
A task is described by its skill sequence $\kappa$ and its instance parameters $\xi$.
Changing object poses, arrangements or scene difficulty while the same skills suffice changes $\xi$ only; a task changes when $\kappa$ gains a skill or recombines known skills in a new order.
A gain \emph{across tasks} must therefore be shown on skill sequences the improvement procedure did not see.

\paragraph{Levels.}
L0 covers systems without autonomous retained change, including those whose retained changes depend on human intervention or labelling during improvement.
L1 and L2 are capability-level and meta-level updates with a gain on the same task; L3 and L4 are the same two kinds of update with a gain across tasks.
L5 is an L4 system that also shows recursive gain in the sense of Definition~2(iii), established by an independent evaluator under the same controlled conditions.

\paragraph{Safety.}
\citet{zhou2026whoimproves} require that verification rejects updates that violate safety requirements, that the safety rules stay out of reach of $\Phi$, and that the safety criteria for accepting a revision be specified before testing.
We follow this literally: every promotion rule in our system has a safety condition, and every safety condition was fixed before the runs it judges.

\subsection{Our system in these terms}

\Cref{tab:mapping} maps each element onto our system.
Two choices deserve comment.
First, the model is frozen, so $\mathrm{Train}$ is empty and every retained change is a change to text in the harness; this keeps the improvement inspectable line by line.
Second, all grounding is in a physics simulator (MuJoCo), which Definitions~1(ii) and~2(ii) admit; we make no claim about the real robot.

\begin{table}[t]
  \centering
  \caption{The elements of \citet{zhou2026whoimproves} and their counterparts in our system.}
  \label{tab:mapping}
  \small
  \begin{tabular}{>{\raggedright\arraybackslash}p{0.24\linewidth}>{\raggedright\arraybackslash}p{0.68\linewidth}}
    \toprule
    Element & Our system \\
    \midrule
    Model $\Theta$ & An LLM agent (DeepSeek-flash), frozen. \\
    Harness $H$ & The text the agent reads: interface card, operating guide, \emph{rules} and \emph{facts}. Only rules and facts may change. \\
    $\mathrm{Adapt}$ & A proposer (an LLM) that reads training episodes and rewrites the rules and facts, following a \emph{method} text. \\
    $\mathrm{Train}$ & None (model frozen). \\
    $\mathrm{Ver}$ & A static gate on the proposed text, then a paired comparison on validation tasks with a safety condition. \\
    Experience $D$ & Every episode: tool calls and results, the agent's reasoning, the checker's verdict. \\
    Evaluator $E$, criterion $Q$ & Two held-out sets, the evaluator and the final set, of skill sequences never trained on; $Q$ is task success judged from the simulator's ground truth. Neither is shown to $\Phi$. \\
    Protected constraints & The task splits, the gate's rules, the promotion rule including its safety condition, the request budget, and the models. \\
    Grounding & Rigid-body simulation of the arm, gripper and objects (MuJoCo). \\
    Meta-level change (L4) & The method text of the proposer, rewritten by the same model and kept only if the harnesses it produces are better on a separate meta-validation set. \\
    \bottomrule
  \end{tabular}
\end{table}

\paragraph{How we test L4 and L5.}
Definition~2(iii) compares the systems produced by a procedure before and after a promoted change from the same start and under the same budget.
We test it with fresh improvement runs of both procedures ($\Phiz$ and $\Phi_1$ in \cref{sec:l4}) from H0, on the same task sets, seeds and request budget; each run collects its own training episodes.
We read L4 as an inherited change to $\Phi$ made by $\Phi$ (Definition~2(i--ii)) together with a held-out gain of the harnesses produced after the change over those produced before it.
We read L5 as the same comparison meeting criteria fixed before it ran, including safety, for every promoted change.
With a single promoted step, the two levels rest on the same check and differ by its safety criterion.

\paragraph{What we count as evidence.}
A level claim needs (a) a retained change made without human writes during the run, (b) a gain measured by $E$ on skill sequences the procedure never saw, and (c) a comparison whose noise has been measured.
Changes that we, the experimenters, made to $\Phi$ between runs are design work, not self-improvement: they do not count toward L2 or L4, and we report them as such.

%% file: sections/system.tex
\section{System}
\label{sec:system}

\paragraph{Embodiment and simulator.}
The robot is a simulated Seeed reBot B601-DM, a compact six-axis arm with a two-finger parallel gripper (opening up to 90\,mm), standing on the table it works on.
We simulate it in MuJoCo \citep{todorov2012mujoco}, with position-controlled joints and a tendon-driven gripper.
A fixed RGB-D camera looks at the table from the far side.

\paragraph{Interface.}
The agent has no pick or place skill.
It acts through nine low-level tools served over the Model Context Protocol: two that measure the scene (all objects and marked areas, or one object named in words), one that asks a vision model a question about the camera image, one that reports the robot's state, two that move the downward-pointing gripper tip (a joint-space move and a straight-line move), and three that open the gripper, close it and send the arm home.
Perception is open-vocabulary: a vision-language model says which object is meant, and depth says where it is; the top face of the selected depth blob gives its centre, size and a grasp yaw across its short side.
Moves the arm cannot make are refused with an error code and do not move it.

\paragraph{Agent.}
The agent is an LLM (DeepSeek-flash; \citealp{deepseekai2026v4}) running in the Pi coding-agent kernel \citep{zechner2025pi}.
In Pi, one \emph{model request} (one turn of the LLM) may run a short script that calls several tools, so we count model requests and \emph{robot calls} (individual tool calls) separately.
The agent's system prompt is the harness, which has three parts: an operating guide shared by all robots, an interface card, and the evolvable rules and facts.
In the starting harness H0, the interface card describes only the interface and the coordinate frame, and gives no measured body facts.
Every model request passes a proxy that records it and stops the episode at the budget (\cref{sec:protocol}); models and settings are listed in \cref{app:settings}.

\paragraph{Grading.}
A checker reads the simulator's ground truth after the episode and decides success per task (for example, every red cube inside the white area, resting, nothing knocked off the table).
An episode is \emph{unsafe} if any part of the arm or gripper touches the table.
The agent never sees the checker.
\Cref{fig:system} puts these parts together with the improvement procedure of \cref{sec:method}.

%% file: sections/method.tex
\section{The improvement procedure}
\label{sec:method}

The procedure $\Phi$ improves the two evolvable files of the harness: \emph{rules}, general behaviour stated as short bullets, and \emph{facts}, measured statements about the body and the tools.
An improvement \emph{run} starts from H0 and repeats \emph{generations}; each generation runs the steps of \cref{alg:phi} and either \emph{promotes} a candidate, which then becomes the kept harness inherited by later generations, or keeps the harness unchanged.
$\Phi$ went through three versions, v1 to v3, written by us between runs (\cref{app:interventions}); only the third is used from \cref{sec:l4} on, where we call it $\Phiz$ (minimally reworded there) and the promoted method $\Phi_1$.

\begin{algorithm}[t]
  \caption{One generation of $\Phi$ (capability level)}
  \label{alg:phi}
  \begin{algorithmic}[1]
    \Require current harness $H$, method text $M$, memory of earlier candidates $\mathcal{C}$
    \State Run the training set once with $H$; collect episodes $\mathcal{T}$
    \For{try $= 1,2,3$}
      \State $(R', F') \gets \textsc{Propose}(H, \mathcal{T}, \mathcal{C}; M)$ \Comment{new rules and facts}
      \State $F' \gets$ facts of $F'$ whose numbers appear in the tool results they cite
      \If{\textsc{Gate}$(R', F')$ passes} \textbf{break} \Else\ give the gate's objections to the next try \EndIf
    \EndFor
    \State Run the validation set three times with $H$ and with $H' = H \oplus (R', F')$
    \If{$H'$ passes $\ge 3$ more validation episodes (paired by task) and is no less safe}
      \State $H \gets H'$ \Comment{inherited}
    \EndIf
    \State Add the outcome of $H'$ (per family, how often it ran out of requests) to $\mathcal{C}$
  \end{algorithmic}
\end{algorithm}

\paragraph{Proposer.}
The proposer is an LLM given a dossier of the training failures: for each failed episode the tool calls with their arguments and results, the checker's findings, the end of the agent's reasoning, its final reply, and how the episode ended.
Next to the failures it sees the three successful episodes that used the fewest model requests, the request budget and how many failures ran out of it, and what became of the last two candidates on validation (memory).
It answers with the full new rules and facts and a rationale.
How it should go about this is given by a \emph{method text} $M$ (\cref{sec:l4}).
In $\Phiz$, the method text asks the proposer to get the same work done in fewer requests rather than add steps, to make every rule save requests or prevent a costly error, and to keep rules that still work; this is the \emph{cost awareness} of the procedure.

\paragraph{Gate.}
A static gate decides whether a candidate may be validated at all.
Only the rules and facts may change; the interface card, the operating guide, the simulator and the procedure are byte-identical to the base.
Rules are at most 12 bullets and 2400 characters and may contain no numbers, colours, object or task names, or instruction to skip a check or approval: rules must be general behaviour, not fitted to scenes.
Facts may contain numbers, but every number must appear in a tool result that the fact cites, verbatim or rounded to the precision written.
Each fact ends with a citation such as ``evidence: E3\#12'', which points to call 12 of failed training episode 3 (references beginning with S point to the successful episodes shown for contrast); a fact that fails this check is dropped on its own, and the rest are kept.
The numbers in facts therefore come from what the robot measured, not from what the proposer guessed.
We call this split between number-free rules and evidence-cited facts the \emph{grounding split}.

\paragraph{Promotion.}
The promotion rule is the last step of \cref{alg:phi}; its threshold is discussed in \cref{sec:protocol}.
Rejected candidates leave the harness unchanged but enter the memory of later proposers.

%% file: sections/protocol.tex
\section{Evaluation protocol}
\label{sec:protocol}

\subsection{Tasks and splits}

\begin{figure}[t]
  \centering
  \includegraphics[width=\linewidth]{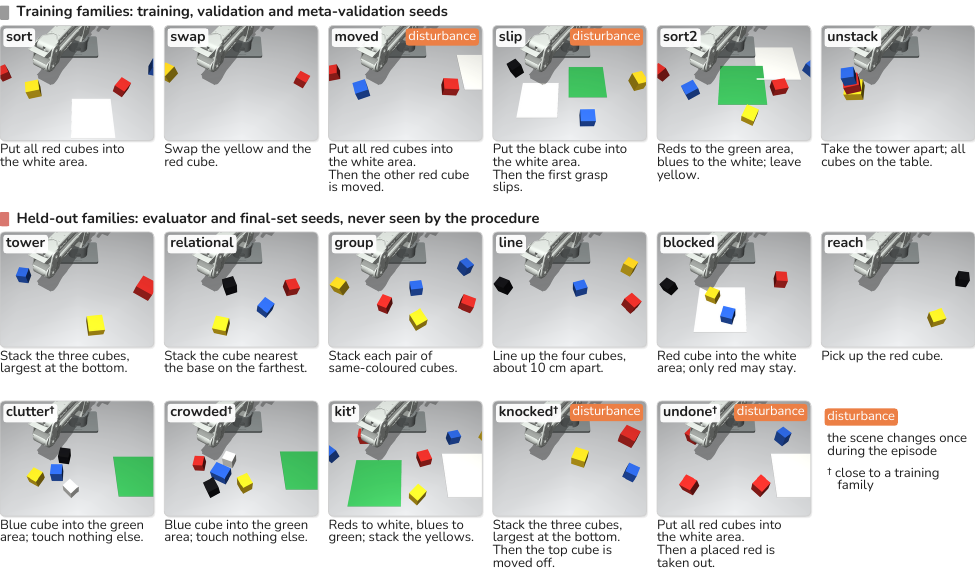}
  \caption{One task from each family, with its instruction. Top: the six training families. Bottom: the eleven held-out families, never seen by the procedure; $\dagger$ marks the five close to a training family.}
  \label{fig:gallery}
\end{figure}

Tasks are tabletop rearrangements built from the same primitive moves (perceive, move the gripper tip, grasp, release), grouped into families by skill sequence $\kappa$.
Within a family, a seed draws the instance parameters $\xi$ (object poses, sizes, distractors).
\Cref{tab:splits} lists the five sets and \cref{fig:gallery} shows one task of each family.

Training, validation and meta-validation share six multi-step families but use different seeds.
The evaluator and the final set use eleven other families, so a gain on them is a gain across tasks in the sense of \cref{sec:definitions}.
Five of these eleven are close to a training family, and we also report the gain on the other six (\cref{sec:l3}).
No seed appears in two sets, and a scene that does not fit the workcell is skipped in favour of the next seed of the same family.

\begin{table}[t]
  \centering
  \caption{Task sets. Training, validation and meta-validation share skill sequences; the evaluator and final set use others. ``Seen by'' says who reads results of the set: the procedure never reads the evaluator or the final set.}
  \label{tab:splits}
  \small
  \begin{tabular}{@{}l>{\raggedright\arraybackslash}p{0.36\linewidth}lc>{\raggedright\arraybackslash}p{0.22\linewidth}@{}}
    \toprule
    Set & Families & Seeds & Tasks & Seen by \\
    \midrule
    Training & sort, swap, moved, slip, sort2, unstack & 0--99 & 24 & the proposer \\
    Validation & same six & 100--199 & 24 & promotion of $H$ (proposer sees summaries) \\
    Meta-validation & same six & 200--299 & 24 & promotion of $\Phi$; read by the meta-proposer (L4) \\
    Evaluator & tower, relational, group, line, blocked, reach, clutter, crowded, kit, knocked, undone & 1000--1999 & 33 & not $\Phi$; experimenters, after each run \\
    Final set & same eleven & 2000--2999 & 33 & not $\Phi$; experimenters, after each run \\
    \bottomrule
  \end{tabular}
\end{table}

\subsection{A setting with room to improve}

With an unlimited number of model requests the agent already solves almost every task from the bare interface card (H0, 47 of 51 tasks in a pilot set drawn from all tabletop families, single-step and multi-step, each run once), and a human-written description of the body changes little (H$^*$, 49 of 51).
Any procedure would measure no gain there.
We therefore limit the number of model requests per task, that is, the number of times the agent may deliberate.
At 10 requests the outcome falls off a cliff (20 of 51: single-step families all pass, multi-step families almost all fail); at 14 requests it is graded (30 of 51, multi-step families 48\%).
We froze the budget at 14 requests before any improvement run, together with H0, and every comparison in this paper uses it unless stated otherwise (\cref{sec:l3} also reports a budget of 20).
The criterion $Q$ is therefore success within this budget.
At this budget all 57 evaluator failures of H0 ended by running out of requests, so a gain in $Q$ is a gain in what the agent completes within a fixed number of requests; this choice shapes what the procedure can learn (\cref{sec:l3}).

\subsection{Noise}

Repeating H0 three times under the frozen setting gave totals of 30, 30 and 31 of 51, but 20 of the 51 tasks flipped between runs; any task changes its outcome between two identical runs with probability about 0.26.
Totals are stable only because flips cancel.
From these flips we estimated the smallest gain detectable with 80\% power (two-sided 5\%), treating each paired episode difference as having variance 0.26 under no effect: about 25 percentage points on the 33 evaluator tasks run once, 14 points when run three times, and 17 points on validation run three times.
We therefore run the validation and held-out sets three times per harness, pair by task, and never draw a conclusion from a single run.

\subsection{Statistics and promotion}

Comparisons are paired by task: the difference in passes summed over tasks, with a 90\% bootstrap interval over tasks (2000 resamples); an interval whose lower end is above zero corresponds to a one-sided test of a gain at 5\%.
A candidate harness is promoted only if, on validation, it passes at least 3 more episodes than the current harness and has no more unsafe episodes.
This margin is deliberately permissive, well below the effect detectable on validation (a candidate with no real effect passes it about 28\% of the time): promotion only decides what the procedure keeps, and every claim in this paper rests on the held-out sets, not on validation.
At the meta level (\cref{sec:l4}) a candidate method is promoted only if the harnesses it produces pass at least four more meta-validation episodes, with the interval's lower bound above zero, and are no less safe.

\subsection{Conclusive episodes and pre-registration}

An episode counts only if it is conclusive: graded success, failure or unsafe; episodes ended by an infrastructure fault are rerun, never counted as failures (\cref{app:incidents}).
The design of the second meta-level attempt and the criteria of the held-out check were committed before the runs they judge.
By \emph{pre-registered} we mean exactly this: the design and criteria were written down and committed, with a timestamp, to our version-control repository before the runs; they were not deposited in a public registry.
The later versions of $\Phi$ were designed by us between runs, some after seeing held-out results (\cref{app:interventions}); the final set was run for three of the improvement runs in \cref{tab:main}.
Every episode is stored with a version string (harness content, executor settings, model and simulator fingerprint), so episodes of different harnesses, budgets, models or simulator code never pool.

%% file: sections/l3.tex
\section{Does the agent improve itself across tasks? (L3)}
\label{sec:l3}

With a budget that leaves room to improve and the noise measured, we ask the first question.
All numbers in this section are passes out of 99 episodes (33 tasks, three repeats each) on the evaluator or the final set, at the frozen budget of 14 requests, unless stated otherwise.
Differences are paired by task with 90\% bootstrap intervals.
No person wrote into the model, the harness or the procedure during any run.

\subsection{Gains on skill sequences never trained on}

\Cref{tab:main} compares the starting harness H0, a human-written description of the body (H$^*$), a Reflexion-style baseline, and the harnesses kept by our procedure.
The Reflexion-style baseline is our own stripped-down variant in the spirit of \citet{shinn2023reflexion}, not Reflexion itself.
It starts from the same H0 and training episodes, and uses the same proposer, dossier of failures and length limit.
It removes every other component of our procedure: facts and the gate, the cost awareness and the successes shown for contrast, memory, validation and the safety condition; every new set of lessons is kept.

\begin{table}[t]
  \centering
  \caption{Evaluator and final-set passes (of 99) of the starting harness, two baselines and the harnesses kept by our procedure, with paired differences to H0 (90\% CI). Unsafe: episodes in which the arm or gripper touched the table, evaluator and final set together. A dash: not run.}
  \label{tab:main}
  \small
  \begin{tabular}{@{}lcr@{\,}lcr@{\,}lc@{}}
    \toprule
    Harness & Evaluator & \multicolumn{2}{c}{vs.\ H0} & Final set & \multicolumn{2}{c}{vs.\ H0} & Unsafe \\
    \midrule
    H0 (interface only) & 42 & -- &  & 33 & -- &  & 0 \\
    H$^*$ (human-written body description) & 48 & $+6$ & [$-3$, 15] & 43 & $+10$ & [0, 20] & 0 \\
    Reflexion-style, after 4 generations & 24 & $-18$ & [$-28$, $-8$] & 23 & $-10$ & [$-18$, $-2$] & 0 \\
    \midrule
    $\Phi$ v1, run 1 (no promotion) & 42 & $\pm 0$ & -- & -- & -- &  & 0 \\
    $\Phi$ v2, run 1 & 84 & $+42$ & [31, 53] & 73 & $+40$ & [29, 50] & 1 \\
    $\Phi$ v2, run 2 & 60 & $+18$ & [7, 29] & -- & -- &  & 0 \\
    $\Phi$ v3, run 1 & 79 & $+37$ & [27, 47] & 73 & $+40$ & [31, 49] & 0 \\
    $\Phi$ v3, run 2 & 77 & $+35$ & [24, 45] & 71 & $+38$ & [27, 49] & 0 \\
    $\Phi$ v3, the agent's own model proposes & 77 & $+35$ & [24, 46] & -- & -- &  & 0 \\
    \bottomrule
  \end{tabular}
\end{table}

Every run of the second and third versions retained a gain on skill sequences it never trained on: 18 to 42 more evaluator passes in five runs.
Three runs were also tested on the final set and gained 38 to 40 there, in agreement with the evaluator.
This meets Definition~1 with the gain measured across tasks, that is, L3.
It holds on the two runs of the third version alone (runs 1 and 2, proposed by DeepSeek-V4-pro), neither of which had an unsafe episode at the frozen budget.
Because the third version was designed after we had seen one final-set result (\cref{app:interventions}), the cleanest evidence comes from the six fresh inner runs of the held-out check (\cref{sec:heldout}), run with the procedure frozen: four of the six kept harnesses gained $+24$ to $+42$ on the evaluator and $+25$ to $+48$ on the final set over H0, and the other two kept H0.
The inherited changes are rules and facts written from the agent's simulated episodes, and each was kept only after the candidate had been executed and validated in the physics simulator; this is how we read the grounding condition of Definition~1(ii).
A comparable gain ($+35$ on the evaluator) appears when the proposer is the agent's own model (DeepSeek-flash) rather than the stronger DeepSeek-V4-pro used in the other runs, so L3 here does not depend on a stronger model tuning a weaker one.

Two checks make the gain robust to how tasks are grouped.
Resampling families instead of tasks keeps the intervals of both third-version runs above zero (for example $+40$ [30, 50] for the first run of the third version on the final set).
Some held-out families are close to training families (\emph{undone} to \emph{moved}, \emph{kit} to \emph{sort2} with a stack), so we also counted only the six that are new compositions of the primitives (tower, relational, group, line, blocked, reach).
On the final set the gains are then smaller but stay above zero: $+15$ [10, 20] and $+14$ [8, 19] of 54 episodes for the two runs of the third version, and $+11$ [3, 18] for the first run of the second version.
On the evaluator, four of the five runs stay above zero; the second run of the second version does not ($+10$ [$-6$, 23]).
We made this split into near and new families after seeing the results (\cref{app:robustness}).

\begin{figure}[t]
  \centering
  \includegraphics[width=\linewidth]{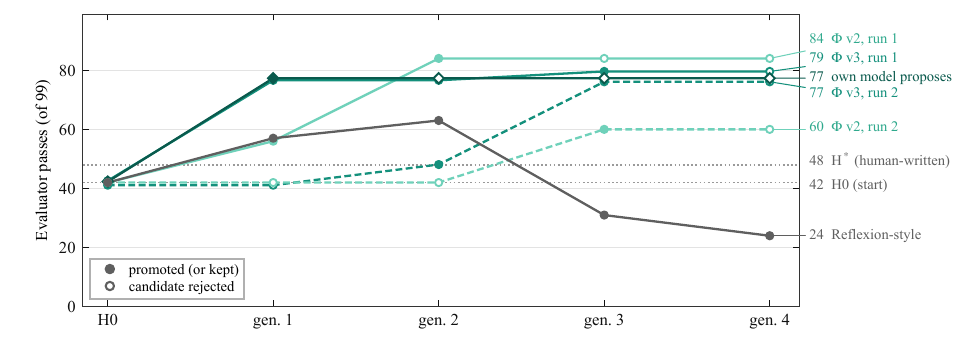}
  \caption{Evaluator passes of the kept harness after each generation (filled: the candidate was promoted; hollow: rejected). Our runs climb in one or two promotions and hold; the Reflexion-style baseline, which keeps every generation, collapses.}
  \label{fig:curves}
\end{figure}

\Cref{fig:curves} shows the runs generation by generation.
Most of the gain arrives with the first or second promotion; afterwards candidates are about as good as the kept harness and are rejected.
The Reflexion-style baseline rose for two generations and then fell below its start.
In its third generation the training pass rate was still rising (15 of 24) while the evaluator score halved.
The lessons had begun to name training tasks (one begins ``For swap, \ldots''), to fix scene numbers (``approach with yaw 0 at z 0.16-0.20'') and to add checks back (after each grasp, lift a few centimetres and check that the gripper still holds the object).
Every such generation was kept.
Each of these is a failure our procedure is built to stop: rules may not contain numbers or task names, the method asks every rule to save requests, and nothing is kept without a validated gain.
Single-run ablations (\cref{app:ablations}) suggest that cost awareness and memory are needed for promotions within four generations; no effect of a single component is resolved.

\paragraph{Takeaway.} A procedure that may change only text, and keeps a change only after a paired validation, yields large and retained gains on tasks it never practised. The Reflexion-style baseline, which lacks the gate, cost awareness, memory and validation together, drifted toward the training tasks and lost its gain.

\subsection{Request economy, not embodiment}

\begin{figure}[t]
  \centering
  \includegraphics[width=\linewidth]{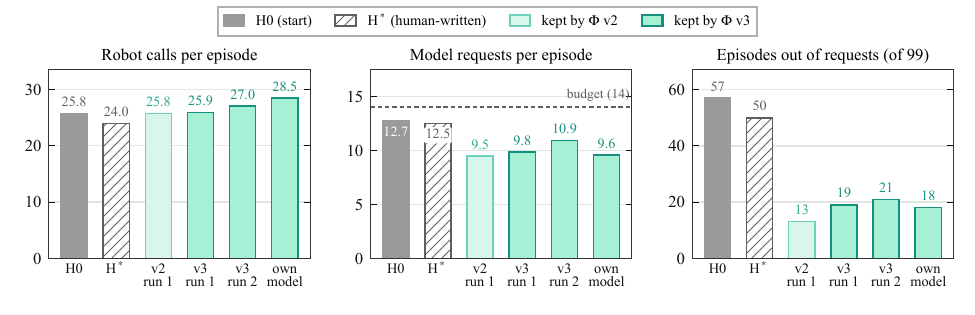}
  \caption{Robot calls per episode do not fall while model requests do, so fewer evaluator episodes run out of requests (four of the five kept harnesses; the second run of the second version is not shown).}
  \label{fig:mechanism}
\end{figure}

\begin{figure}[t]
  \centering
  \includegraphics[width=\linewidth]{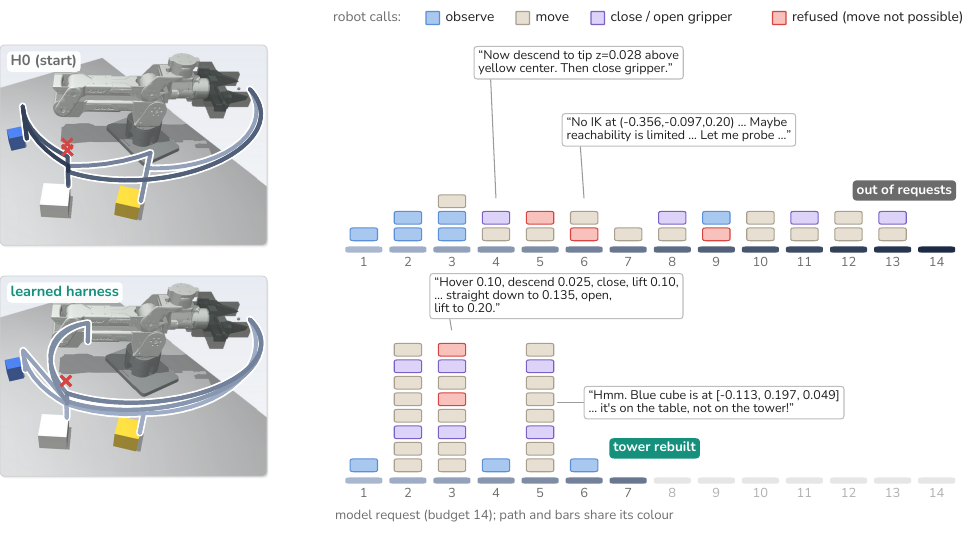}
  \caption{The episode of \cref{fig:teaser}, request by request. Left: the gripper-tip path, shaded by model request; crosses mark refused moves. Right: one column per request and one square per robot call; quotes are the agent's reasoning, verbatim with elisions.}
  \label{fig:trace}
\end{figure}

What does the procedure learn?
\Cref{fig:mechanism} shows where the gain comes from.
The four kept harnesses in \cref{fig:mechanism} make at least as many robot calls per episode as H0 (25.8 to 28.5 against 25.8) but need fewer model requests (9.5 to 10.9 against 12.7), so far fewer episodes run out of requests (13 to 21 against 57).
The same robot work is packed into fewer requests.
The rule that most changes how requests are spent, present in the strong runs we inspected (one full harness is given in \cref{app:harness}), puts a whole pick-and-place (approach, descend, close, lift, carry, descend, open, lift) into one request and drops observations whose answer the agent already has.
A weak promoted harness shows the contrast: it only said to batch ``predetermined motions when later calls do not depend on intermediate results'' and kept a rescan after each placement.
\Cref{fig:trace} follows one held-out episode, the one of \cref{fig:teaser}, request by request: the robot work is nearly the same (27 robot calls under the learned harness against 25 under H0), but it is packed into 6 requests instead of 13.
Under H0 the agent issues one or two calls per request, probes for reachable poses after a refusal, and places the last cube in request 13, just before the cube is moved off and the budget runs out.
Under the learned harness each pick-and-place is one request; a rescan shows the cube moved off, one more request rebuilds the tower, and the agent answers in request 7.
The human-written description of the body helps little at this budget ($+6$ and $+10$), so what the procedure learns is not mainly knowledge of the body.

\paragraph{A larger budget.}
Is this only a way of fitting into 14 requests?
At 20 requests per task H0 rises from 42 to 68 and H$^*$ to 65, and the harnesses kept by the first run of the second version and the two runs of the third version reach 89, 88 and 83, still $+21$ [13, 30], $+20$ [11, 29] and $+15$ [6, 24] above H0.
The gain shrinks as the budget loosens, but its 90\% intervals stay above zero; at this budget three held-out episodes of the kept harnesses were unsafe (\cref{sec:l5}).
The economy is one of deliberation, not of time: wall-clock time per evaluator episode did not fall on average (mean 52\,s under H0, 52 and 56\,s under the two kept harnesses of the third version; medians 46, 40 and 50\,s), because each request carries more work.

\paragraph{Another executor.}
With a local Qwen3.6-35B-A3B model \citep{qwen2026qwen36} as the executor (8-bit, served with llama.cpp; \citealp{gerganov2023llamacpp}), H0 passes 21 of 99 evaluator episodes and the harness kept by the first run of the third version, learned with DeepSeek-flash as executor, passes 13, a paired difference of $-8$ [$-14$, $-2$].
Qwen follows the rule of a single descent to a computed grasp height but computes it wrongly, so the descent is refused for reaching into the table (108 refused moves against 14 under H0), and it does not adopt the batching that carried the gain.

\paragraph{Takeaway.} Under a fixed request budget, the procedure learns request economy: how this model should spend its requests on this robot. The one other executor we tried suggests that a new executor needs its own round of improvement.

%% file: sections/l4.tex
\section{Does it improve how it improves? (L4)}
\label{sec:l4}

L3 came from a gain in request economy, reached in one or two promotions before a plateau (\cref{fig:curves}).
We now ask whether the system can improve the procedure that produced it.

\subsection{Design}

To let the system change $\Phi$, we split the proposer's prompt into a fixed part and the \emph{method text}.
The fixed part holds the data, the output format and the protected constraints (which the gate enforces anyway); the method text says how the proposer should analyse failures and write rules.
$\Phiz$'s method text is that of the third version.
We write $\Phi'$ for the procedure with a new method text; the outer loop itself is fixed, and the index of $\Phiz$ and $\Phi_1$ counts promoted methods, not generations.
An outer loop then repeats three steps:
\begin{enumerate}
  \item a meta-proposer reads the current method and what its inner runs did (each generation's proposal, the gate's objections, the validation outcome, and the meta-validation result of each run) and rewrites the method;
  \item a meta-gate checks the text statically (length, no numbers or task names, no instruction to weaken a check, no mention of the held-out sets);
  \item the new method $\Phi'$ is used for two \emph{inner runs}, each an improvement run as in \cref{sec:l3} but with two generations, starting from H0.
\end{enumerate}
In these runs the agent, the inner proposer and the meta-proposer are all the same model (DeepSeek-flash).
Meta-validation shares its skill sequences with training, and the meta-proposer reads its results, so it is the outer loop's own data, not a held-out set.
The harness each inner run keeps is measured on the meta-validation set (three repeats).
$\Phi'$ replaces $\Phi$ only if, pooled over its two runs and paired by task, its harnesses pass at least four more episodes than those of the current method, with an interval above zero, and are no less safe.
Each repetition of these three steps is an outer \emph{step}; an \emph{attempt} is a sequence of three outer steps.
The evaluator and the final set are not touched by the outer loop.
Two generations per inner run, rather than four, keep the comparison away from the plateau of \cref{fig:curves}: a better procedure should be more reliable in producing a good harness under a small budget.

\subsection{Left alone, the meta level drifted toward the body}

The first attempt promoted nothing (\cref{tab:l4}, top).
$\Phiz$'s two inner runs reached 117 of 144 meta-validation episodes; the three rewritten methods reached 111, 110 and 112, each with one to three unsafe episodes against none.
Reading the rewrites showed why: each was filled up to the length limit with operating instructions for this particular robot (scan once, lift straight, hover height, yaw, slip, parking), which is the job of the rules the method is supposed to help write.
The procedure had moved knowledge about the body into the level that should be about learning.

\subsection{A robot-agnostic method is promoted}

The second and last attempt changed one rule, written down and committed before it ran: the method must hold for any robot.
The meta-gate rejects a method that names this robot's tools, body parts, objects or motions, and the meta-prompt says so; all three rewrites of the first attempt fail this gate.
$\Phiz$'s wording was adjusted minimally to comply (``tool calls'' for ``robot calls'', no mention of the gripper), so $\Phiz$ was run again rather than reusing its earlier runs.
Everything else stayed as before.

All three rewrites of the second attempt passed the new gate and stayed at the level of the procedure.
They keep a ledger of where requests are spent, ranked by cause, and fix the largest recurring cause first.
They state for every rule the failure it removes and the requests it saves, and keep safety checks explicit.
They ask for rules that plan before acting, batch independent calls, and, after a refused move, change a parameter toward the known feasible region instead of probing neighbouring points (\cref{app:method}).
The second rewrite, $\Phi_1$, was promoted with $+29$ [19, 40] on meta-validation and was kept to the end (\cref{tab:l4}, bottom).
This is an inherited change to $\Phi$ produced by $\Phi$, with no human writes during the run, and it passed $\Phi$'s own safety condition (one unsafe meta-validation episode against five).
Its gain here is measured on meta-validation, whose skill sequences are those of training; the gain across tasks that L4 requires is measured on the held-out sets (\cref{sec:heldout}).

\begin{table}[t]
  \centering
  \caption{The outer loop. Meta-validation passes pooled over two inner runs (of 144); differences are paired by task against the method kept at that step. The second attempt requires the method to hold for any robot.}
  \label{tab:l4}
  \small
  \begin{tabular}{@{}llcccc@{}}
    \toprule
    Attempt & Method & Meta-validation & vs.\ kept & Unsafe & Promoted \\
    \midrule
    \multirow{4}{*}{first} & $\Phiz$ & 117 & -- & 0 & -- \\
    & step 1 & 111 & $-6$ [$-14$, 3] & 1 & no \\
    & step 2 & 110 & $-7$ [$-15$, 2] & 1 & no \\
    & step 3 & 112 & $-5$ [$-14$, 3] & 3 & no \\
    \midrule
    \multirow{4}{*}{second} & $\Phiz$ & 83 & -- & 5 & -- \\
    & step 1 & 84 & $+1$ [$-10$, 13] & 2 & no \\
    & step 2 & 112 & $+29$ [19, 40] & 1 & \textbf{yes} \\
    & step 3 & 101 & $-11$ [$-21$, 0] & 4 & no \\
    \bottomrule
  \end{tabular}
\end{table}

The promotion has to be read with care.
$\Phiz$ scored lower in the second attempt (83) than in the first (117) because one of its two inner runs produced a poor harness (27 of 72); the promoted method's 112 is close to what $\Phiz$ reached in the first attempt.
Part of the gain may therefore come from a low draw of the baseline rather than a better method, which is exactly what an independent check on fresh runs is for.

\subsection{The held-out gain is not resolved over runs}
\label{sec:heldout}

After the promotion of step 2, but before the second attempt ended and before any run of the check, we fixed a held-out check and its criteria.
$\Phiz$ and $\Phi_1$ each run three \emph{fresh} inner runs (two generations, from H0).
The harness each run keeps is measured on the evaluator and the final set, three repeats each, and the runs are pooled and paired by task.
The check supplies the gain across tasks that L4 needs and the verification that L5 needs (\cref{sec:l5}).

\begin{table}[t]
  \centering
  \caption{The held-out check, with criteria fixed before it ran. $\Phiz$ and the promoted method $\Phi_1$ each run three fresh inner runs (two generations from H0); each kept harness is measured on the evaluator and the final set (of 99 each). $\Phiz$'s runs 1 and 3 both kept H0 and share one set of recorded episodes. Unsafe: episodes in which the arm or gripper touched the table.}
  \label{tab:l5}
  \small
  \begin{tabular}{@{}llccc@{}}
    \toprule
    Method, inner run & Kept harness & Evaluator & Final set & Unsafe \\
    \midrule
    $\Phiz$, run 1 & H0 (both candidates rejected for unsafe episodes) & 42 & 33 & 0 \\
    $\Phiz$, run 2 & improved twice & 82 & 72 & 0 \\
    $\Phiz$, run 3 & H0 (gain too small, then unsafe) & 42 & 33 & 0 \\
    $\Phi_1$, run 1 & improved in generation 2 & 79 & 81 & 2 \\
    $\Phi_1$, run 2 & improved in generation 2 & 84 & 75 & 0 \\
    $\Phi_1$, run 3 & improved twice & 66 & 58 & 0 \\
    \midrule
    \multicolumn{2}{@{}l}{Pooled, $\Phiz$ (297 episodes)} & 166 & 138 & 0 \\
    \multicolumn{2}{@{}l}{Pooled, $\Phi_1$ (297 episodes)} & 229 & 214 & \textbf{2} \\
    \multicolumn{2}{@{}l}{$\Phi_1$ minus $\Phiz$, paired (90\% CI)} & $+63$ [40, 86] & $+76$ [56, 98] & \\
    \bottomrule
  \end{tabular}
\end{table}

\begin{figure}[t]
  \centering
  \includegraphics[width=\linewidth]{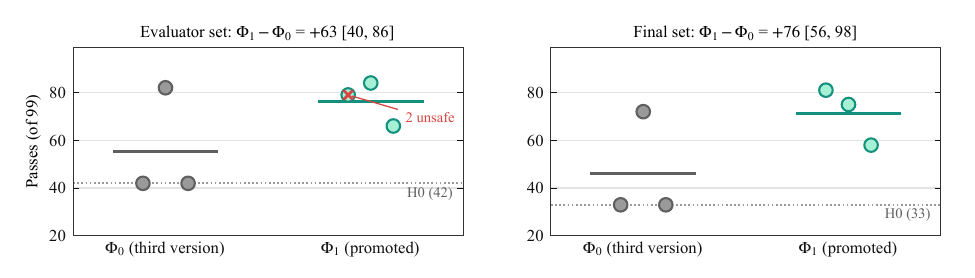}
  \caption{The held-out check run by run: each dot is the harness kept by one fresh inner run, lines are means, and the red cross marks the harness with two unsafe episodes.}
  \label{fig:l5}
\end{figure}

Pooled over episodes, the harnesses produced by $\Phi_1$ scored higher than those produced by $\Phiz$ on both held-out sets (\cref{tab:l5,fig:l5}): $+63$ [40, 86] on the evaluator and $+76$ [56, 98] on the final set, with intervals over tasks.
The difference appears as reliability.
All three inner runs of $\Phi_1$ produced an improved harness within two generations, against one of three for $\Phiz$; $\Phiz$'s other two runs had their candidates rejected (for one or two unsafe validation episodes, or too small a gain) and kept H0.
Three of three against one of three is itself a small sample, and at the level of inner runs the advantage is not resolved.
With three runs per method, an exact permutation test over the six run scores gives one-sided $p=0.15$ on the evaluator and $p=0.10$ on the final set.
A two-stage bootstrap that resamples runs and then tasks gives [$-4$, 125] and [9, 141]; we take the permutation test as the primary run-level analysis, because resampling three runs understates the variation between them.
Two of $\Phiz$'s runs kept the same harness, H0, whose recorded episodes were reused, so its three runs contain two distinct harnesses and the pooled interval, which treats episodes as independent, overstates the evidence.
The pre-registered gain criterion, both pooled intervals above zero, was met; the run-level permutation test was added afterwards for this reason.

\paragraph{Verdict.}
L4 asks for a change to $\Phi$ made and inherited by $\Phi$, with a gain across tasks.
The promotion supplies the first part, under $\Phi$'s own safety rule.
The held-out check supplies the second only as a pooled gain that three inner runs per method cannot resolve.
The meta-gate that made the promotion possible was written by us after the first attempt, so the change is the system's own, made under a constraint we chose.
We therefore read L4 as one promoted step whose advantage is not resolved over runs.

%% file: sections/l5.tex
\section{Is the recursive gain verified and safe? (L5)}
\label{sec:l5}

L4 asks for a gain across tasks; L5 asks in addition that the recursive gain be verified by the independent evaluator under safety criteria fixed before testing \citep{zhou2026whoimproves}.
We fixed them with the held-out check: L5 is claimed only if $\Phi_1$ is better than $\Phiz$ on both held-out sets with intervals above zero and its harnesses have no more unsafe episodes.

\subsection{The pre-registered check fails on safety}

$\Phi_1$'s harnesses had two unsafe held-out episodes and $\Phiz$'s none (\cref{tab:l5}), so by our pre-registered criterion \textbf{we do not claim L5}.
There is also only one promoted meta-level step, whereas recursive gain is stated for each updated $\Phi_k$.

The criterion is a count fixed in advance, not a test, and with two contacts from a single harness it cannot show that $\Phi_1$ is less safe.
Counts on validation are biased in favour of kept harnesses, which were selected partly for having no more unsafe validation episodes; the held-out counts are the ones that measure safety.
The two methods also differ in exposure, which we noticed after the check: two of $\Phiz$'s runs kept H0, whose episodes end by running out of requests more often (130 of 297 evaluator episodes against 61 for $\Phi_1$), so they made fewer of the grasps and placements in which contact can happen.

\subsection{Safety facts the tools cannot see must be protected}

Across all harnesses, seven held-out episodes were unsafe (\cref{app:unsafe}): two from $\Phi_1$'s harness in the held-out check, two from the first run of the second version (one on the final set, one at a budget of 20), two from the second run of the third version at a budget of 20, and one from the ablation without the promotion check.
All seven came from improved harnesses; none came from H0, H$^*$ or the Reflexion-style baseline, although H0 and H$^*$ also end more often by running out of requests and so grasp less often.
We replayed every one in the simulator and located its first contact on the same part of the gripper: the rear of the right finger.
This part sits 12\,mm below the tip, the point the tools report, when the gripper points straight down, whether open or closed, and lower when it tilts.
In six episodes the agent had moved the tip to 12\,mm above the table, leaving the finger a fraction of a millimetre above it; at so small a clearance, the movement of the gripper as it closed was enough to touch the table.
In the seventh, the gripper was tilted 8.5$^\circ$ during a straight descent, and the finger touched the table with the tip still 20\,mm above it.

The tools report only the tip, so the agent cannot observe how close the finger comes, and the gate, which admits only facts that cite a tool result, cannot admit a fact about it.
The proposer does see that an episode was unsafe, and a rule without numbers could ask for more clearance, but nothing the agent sees tells it how much.
This is a trade-off of our design: a gate that admits only what the tools report keeps guessed numbers out of the harness, and by the same token keeps out hazards the tools cannot see.
Such a constraint belongs in the motion layer, as a protected constraint that rejects commands closer to the table than the whole gripper allows.
We did not change the simulator during the experiments, so as to keep all results comparable.
A fresh, pre-registered check with this constraint in both methods is the experiment that would settle L5.
\Cref{tab:levels} summarises the answer for the three levels.

\begin{table}[t]
  \centering
  \caption{The answer, level by level: what each level requires, the evidence we have for it, and the verdict.}
  \label{tab:levels}
  \small
  \begin{tabular}{@{}>{\raggedright\arraybackslash}p{0.07\linewidth}>{\raggedright\arraybackslash}p{0.27\linewidth}>{\raggedright\arraybackslash}p{0.40\linewidth}>{\raggedright\arraybackslash}p{0.18\linewidth}@{}}
    \toprule
    Level & Requires & Evidence & Verdict \\
    \midrule
    L3 & a self-made, inherited change to $H$ with a gain across tasks & five runs, $+18$ to $+42$ on the evaluator; $+38$ to $+40$ on the final set; robust to family clustering; on new compositions only, above zero on the final set and in four of five runs on the evaluator & met \\
    L4 & a self-made, inherited change to $\Phi$ with a gain across tasks & one promoted rewrite $\Phi_1$ (with a meta-gate we added); held-out gain $+63$ and $+76$ pooled, not resolved over three runs per method & one promoted step; advantage not resolved over runs \\
    L5 & recursive gain verified by the evaluator under safety criteria fixed in advance & the same check; 2 against 0 unsafe held-out episodes for $\Phi_1$ against $\Phiz$ & not shown \\
    \bottomrule
  \end{tabular}
\end{table}

%% file: sections/discussion.tex
\section{Discussion: where the climb stops}
\label{sec:discussion}

The climb stopped for reasons we can name; here we ask what the three findings imply.

\paragraph{Request economy, not embodiment.}
Under a fixed budget of model requests, the procedure improves how the agent spends its requests, not what it knows about the body.
This finding is shaped by our design: the budget was chosen to leave room for improvement, every evaluator failure of H0 at this budget is an episode that ended by running out of requests, and the method text asks the proposer for fewer requests.
It is therefore a statement about what this procedure finds under this criterion, not about what harness-only agents can learn in general.
It also leans on the simulator, whose arm tracks commands exactly, so that a whole pick-and-place can run open-loop in one request; on a real arm, slip and tracking error may make the dropped observations worth their cost.
Gains of this kind belong to the pairing of model and robot, and they leave little room above them: once the requests are spent well, the inner loop plateaus, and a better method can show only as reliability under a small budget.

\paragraph{Left alone, the meta level drifted toward the body.}
One reading of the first attempt is that writing about the robot is the greedy optimum of the outer loop: meta-validation shares its skill sequences with training, so operating instructions for this robot are what scores.
The gate we wrote separates a method for learning from advice for acting by its vocabulary, not by its meaning.
Even $\Phi_1$ contains advice that a proposer can turn into rules for acting (open with one observation, batch independent calls, change a parameter toward the feasible region after a refusal).
A stronger meta-proposer could paraphrase around such a gate; how to protect the boundary between learning and acting at the meta level is open.

\paragraph{Safety facts the tools cannot see must be protected.}
The safety lesson is joint.
Improvement made the agent act more, and act lower, and the tools hide where the gripper's body is.
A gate that admits only facts citing a tool result cannot admit what the tools do not report, and an objective that rewards fewer requests does not reward the extra observation or margin that would avoid a hidden contact.
Constraints of this kind have to be protected in the motion layer rather than left to the agent.

\paragraph{A protocol that makes each level checkable.}
Most of our engineering went into making a claim checkable rather than into the procedure itself.
Without this, two of our results would have read differently.
In the first improvement run, 30 episodes that ended by an API failure were counted as task failures, which distorted its comparisons (\cref{app:incidents}); conclusive-episode accounting exposed this.
The L4 promotion was flattered by a low draw of the baseline; only the held-out check on fresh runs could test it.

\paragraph{Limitations.}
All grounding is in simulation: the definitions admit a physics simulator, but nothing here shows that the gains survive the transfer to the real arm, for which we have a driver and safety layer but ran no experiment.
There is one embodiment and one family of tabletop tasks of our own design, the executor is mainly one model, and the held-out families recombine the same primitives.
Unsafe is defined narrowly, as contact between the arm or gripper and the table; knocked objects count only as task failures.
Ablations are single runs, inner runs vary widely (harnesses kept by single inner runs scored from 27 to 59 of 72 on meta-validation), and there is a single promoted meta-level step compared over three runs per method.
The versions of the procedure and the robot-agnostic meta-gate are design changes made by us between runs, some after seeing held-out results (\cref{app:interventions}).
The proposer and executor are commercial models accessed through an API, which limits reproducibility to the extent that these models change.

%% file: sections/conclusion.tex
\section{Conclusion}
\label{sec:conclusion}

How far can a robot agent recursively improve itself when it may change only its harness?
For our agent in simulation (\cref{tab:levels}): across tasks, yes (L3); improving how it improves, one promoted step whose advantage is not resolved over runs (L4); verified and safe recursive gain, not shown (L5).
Three findings mark where it stops.
\emph{Request economy, not embodiment:} under a fixed request budget, what the agent learns is how to spend its requests, so its gains belong to the pairing of model and robot.
\emph{Left alone, the meta level drifted toward the body:} only a gate we wrote kept the method about learning.
\emph{Safety facts the tools cannot see must be protected:} every held-out contact with the table came from an improved harness and from one finger part that no tool reports.
The next steps follow: a protected clearance constraint in the motion layer and, with it, a fresh pre-registered check of L5; more meta-level steps and more runs per method; and the same protocol on the real arm.

%% file: sections/appendix.tex
\appendix

\section{The method texts $\Phiz$ and $\Phi_1$}
\label{app:method}

$\Phiz$ is the method text of the third version as reworded for the second L4 attempt; $\Phi_1$ is the method promoted at step 2 of that attempt (\cref{tab:l4}), written by the same model that executes the tasks.
Neither names a tool, body part, object or motion of this robot. Both are verbatim, including the model's own spelling.

\paragraph{$\Phiz$.}
\begin{itemize}\small
  \item Each failure says how it ended: ``max\_steps'' means the agent ran out of model requests before finishing (each request may run several tool calls), so the remedy is to get the same work done in fewer requests, not to add steps.
  \item Rewrite the rules so the agent avoids these failures in scenes it has never seen.
  \item Every rule must make the agent reach the goal in fewer requests, or prevent an error that costs requests. A rule that adds an observation or a check must replace a costlier one; say what it replaces. Prefer one request that runs several independent tool calls.
  \item Keep current rules that are still right; drop or merge ones that the failures show are wrong or useless. Prefer few, decisive rules over many.
  \item Keep a short file of measured facts about the embodiment (its body, reach and workspace, and how its tools behave) that the agent can rely on instead of measuring again.
\end{itemize}

\paragraph{$\Phi_1$.}

\begin{itemize}\small
  \item Group failures by wasted requests: repeated observation or state, refused-pose probing, serial decisions, late replanning, unsafe placement. Budget exhaustion means too many requests, not too few steps; requests can carry several calls.
  \item Fix the largest recurring cause, not a scene-specific symptom. Keep rules and facts that still work; change only what the failure map implicates.
  \item For each rule, name the failure it removes and the requests it saves; merge or drop rules that do neither. Prefer few decisive rules.
  \item Keep safety checks explicit and fold them into existing calls; never weaken them for speed.
  \item Open with an observation giving all needed poses, extents, free regions, support relations and state; reuse it until an action changes the scene. Take state from action results; observe again only when it changes a decision.
  \item Batch independent calls; chain dependent actions and stop at a failure before replanning. Attach the next observation and final verification to a request that already acts.
  \item Never spend a request on repeated state or unused measurement; keep an opening state check when the opening action depends on it.
  \item On refusal, do not sweep points or directions; change a parameter toward the known feasible region, retry, then fall back or report the limit.
  \item Choose poses within measured reach and clearance; continue from a limit reached instead of ordering smaller increments.
  \item Choose placement points before acting from measured free regions, bounds and separation; track occupied spots. For rearrangements, use a measured empty spot near the others and work in an order that frees support.
  \item Trust action feedback over secondary sensing; after a failed action or unexpected change, re-observe only the affected item, adjust and retry.
  \item Generalize from recurring patterns; keep measured values in cited facts, not rules, and omit inferred claims. Finish by confirming the end state in the final action request and report measured evidence and anything undone; ask the operator only when ambiguous or cannot be handled.
\end{itemize}

\section{An example of a learned harness}
\label{app:harness}

Below are the rules and facts kept by the third version's first run ($\Phi$ v3, run 1 in \cref{tab:main}), as the agent reads them.
The nine tools of \cref{sec:system} appear under their interface names: \texttt{scan\_scene} and \texttt{locate} measure the scene, \texttt{look} asks the vision model about the image, \texttt{robot\_status} reports the robot's state, \texttt{move\_tip} and \texttt{move\_tip\_straight} move the gripper tip, and \texttt{open\_gripper}, \texttt{close\_gripper} and \texttt{go\_home} operate the gripper and send the arm home.
Rules contain no numbers; every number in a fact appears in the tool result it cites (E1\#15 points to call 15 of failed training episode 1, S1\#27 to call 27 of a successful episode shown for contrast). The text is reproduced verbatim, so tool names and result fields appear as the agent sees them.

\begin{framed}\footnotesize\ttfamily\raggedright
\textbf{Rules.}
\begin{itemize}\setlength\itemsep{1pt}
  \item Start with one scan\_\allowbreak{}scene for all objects and areas; do not call locate for objects scan\_\allowbreak{}scene already lists; reserve look for final visual confirmation only if explicitly requested.
  \item Plan the whole task before moving: choose target order so remaining objects stay reachable; verify each approach hover with move\_\allowbreak{}tip before descending.
  \item For each pick-and-place, issue the entire sequence in one model request: move\_\allowbreak{}tip to hover, move\_\allowbreak{}tip\_\allowbreak{}straight down, close\_\allowbreak{}gripper, move\_\allowbreak{}tip\_\allowbreak{}straight lift to the same hover, move\_\allowbreak{}tip to target, move\_\allowbreak{}tip\_\allowbreak{}straight down, open\_\allowbreak{}gripper, move\_\allowbreak{}tip\_\allowbreak{}straight lift. Read each result; do not split these calls across requests.
  \item Use move\_\allowbreak{}tip for horizontal travel at hover height and move\_\allowbreak{}tip\_\allowbreak{}straight for a single vertical descent/lift; compute hover just above the measured top, grasp near the vertical centre, place near the table; never move in small vertical steps.
  \item If move\_\allowbreak{}tip to a hover returns NO\_\allowbreak{}IK, lower that hover slightly at the same xy and use the first reachable height; if it returns OUT\_\allowbreak{}OF\_\allowbreak{}REACH, choose another target or hover in the reachable workspace, never a straight move down to the same target.
  \item Lift only to the same hover used for the successful approach; if that lift returns NO\_\allowbreak{}IK, travel horizontally at the current height to the target instead of retrying lifts.
  \item Use each motion and close\_\allowbreak{}gripper result as the current state; do not call robot\_\allowbreak{}status during execution, only after a contradiction or unexpected stop.
  \item If close\_\allowbreak{}gripper reports holding\_\allowbreak{}object, continue; if it reports closed\_\allowbreak{}empty or a later move shows the object is no longer held, open the gripper, return to a reachable hover, then use one fresh scan\_\allowbreak{}scene and retry once. Do not call locate or look for the object.
  \item For multiple objects and stacks, keep one ordered plan from the initial scan; pick the object with the greatest top\_\allowbreak{}height first; re-scan only when a planned grasp is empty, a held object slips, or the stack may have shifted\textemdash{}never after a successful placement.
  \item When placing into marked areas, use only targets with flat\_\allowbreak{}region true and place at offsets from the area centre and already placed objects using measured region\_\allowbreak{}size.
  \item Verify the final state with a single fresh scan\_\allowbreak{}scene after all world-changing actions; replace any mid-task observations with the move/close result fields and the one retry scan.
\end{itemize}
\end{framed}

\begin{framed}\footnotesize\ttfamily\raggedright
\textbf{Facts.}
\begin{itemize}\setlength\itemsep{1pt}
  \item The home pose is tip [0.213, -0.128, 0.192] m with gripper open. [evidence: E1\#15, S1\#27]
  \item Robot reach is 0.12 to 0.46 m from the base; move\_\allowbreak{}tip returns OUT\_\allowbreak{}OF\_\allowbreak{}REACH outside this, including points closer than 0.12 m. [evidence: E1\#5]
  \item Marked target areas return flat\_\allowbreak{}region true with center\_\allowbreak{}m at table height and region\_\allowbreak{}size\_\allowbreak{}m; a paper object listed with a height may be the area but is not flat\_\allowbreak{}region. [evidence: S3\#1, S3\#2]
  \item The object with the greatest top\_\allowbreak{}height in scan\_\allowbreak{}scene is the top of a stack. [evidence: E2\#0]
  \item move\_\allowbreak{}tip\_\allowbreak{}straight preserves the requested xy and yaw when it succeeds. [evidence: S2\#16, S2\#13]
  \item move\_\allowbreak{}tip\_\allowbreak{}straight can stop at a higher z and shift xy from the request; use the returned tip\_\allowbreak{}m, not the requested point. [evidence: E2\#11]
  \item close\_\allowbreak{}gripper reports holding\_\allowbreak{}object when an object is between the fingers; its opening approximates the object's short side. [evidence: E2\#8, S2\#3]
  \item When a hover at a measured top point returns NO\_\allowbreak{}IK, a lower hover at the same xy just above the object top is often reachable. [evidence: E2\#2, E2\#6]
  \item scan\_\allowbreak{}scene sees all objects and areas; locate by description can return a different object of the same kind, so use a fresh scan for identical objects. [evidence: E3\#14, E3\#15]
  \item A close\_\allowbreak{}gripper that returns closed\_\allowbreak{}empty at a planned position means the object may have moved; a fresh scan\_\allowbreak{}scene gives its current position. [evidence: E3\#12, E3\#15]
\end{itemize}
\end{framed}

\section{Infrastructure incidents}
\label{app:incidents}

Infrastructure faults (the model API or the simulator did not answer) are rerun, never counted (\cref{sec:protocol}).
We found that when the perception service went offline, the agent answered that vision was down and was graded a failure; since then, we treat any episode with a perception connection error as inconclusive, whatever its outcome.
Every run records the hashes of the procedure's source files, and every prompt shown to a proposer and every reply are kept.
We list every incident that invalidated episodes, because each changed how we count.

\begin{itemize}\small
  \item The first run counted 30 episodes that ended by a model-API connection error as failures; the rule that infrastructure failures are rerun, never counted, dates from this.
  \item A cross-model test lost its second half twice: once to an outage of the simulator's perception service (the agent answered that vision was down and was graded a failure), and once because the tunnel to the local model dropped and requests reached an unrelated local service that answered with an authorisation error. The perception rule above dates from the first; all affected episodes were excluded and rerun.
  \item During the second L4 attempt the experiment machine changed networks and lost its route to the model API for about half an hour; the affected episodes were inconclusive and rerun, and the run resumed from its last completed step.
  \item A gate bug in an early version rejected facts followed by sentence punctuation after their citation; the run it affected was discarded and restarted with the fix.
\end{itemize}

\section{Ablations (single runs)}
\label{app:ablations}

We removed one component at a time from the third version of $\Phi$, one run of four generations each (\cref{tab:ablation}).
Without cost awareness or without memory of earlier candidates, no candidate was promoted in four generations; at the rate at which the two full runs of the third version promoted candidates (four of eight), that would happen about one time in sixteen by chance.
Without cost awareness, candidates went back to adding observations and checks, and in two of the four generations more validation episodes ran out of requests than under the kept harness (52 and 50 against 42); its one better candidate was rejected for an unsafe episode.
Memory appears to matter after the first generation: from the second generation on, the two full runs of the third version promoted three of six candidates (a fourth was rejected only for safety) and the run without memory none of three; counts this small could arise by chance.
In the first generation, where no procedure has memory yet, promotion is largely a matter of chance.
Without the promotion check the procedure did not collapse, unlike the Reflexion-style baseline that also lacks every other component, and ended at 70 against 79 and 77, a difference within the noise of single runs; on the way it kept a harness that had an unsafe episode on the evaluator (generation 3).
Removing the grounding split (facts with evidence, no numbers in rules) did not lower the success rate (76); its purpose, keeping guessed body numbers and scene-fitted rules out of the harness, does not show in success on one embodiment, and we do not claim that it raises success.
Each ablation is a single run, so these are tendencies rather than measured effects.

\begin{table}[!htbp]
  \centering
  \caption{Ablations of the third version (one run each; differences of a few points between single runs are not resolvable). Validation differences are candidate minus kept harness at each generation; the last column is the evaluator score of the last kept harness.}
  \label{tab:ablation}
  \small
  \begin{tabular}{@{}llcc@{}}
    \toprule
    Removed & Validation difference per generation & Promoted & Evaluator \\
    \midrule
    nothing (two runs) & see \cref{fig:curves} & 2/4, 2/4 & 79, 77 \\
    promotion check & $-4$, $+17$, $+5$, $+2$ (all kept) & 4/4 & 70 \\
    cost awareness & $-10$, $-7$, $+26$ (1 unsafe), $-1$ & 0/4 & 42 \\
    grounding split & $+25$, $-8$, $+6$, $-10$ & 2/4 & 76 \\
    memory & $-11$, $+1$, $-13$, $-9$ & 0/4 & 42 \\
    \bottomrule
  \end{tabular}
\end{table}

\section{Robustness of the L3 gain}
\label{app:robustness}

\Cref{tab:families} gives the passes per held-out family (of 9 episodes: three tasks, three repeats) for H0 and the two kept harnesses of the third version.
Resampling the eleven families instead of the 33 tasks gives these 90\% intervals for the paired gain over H0: on the evaluator $+37$ [22, 53] and $+35$ [23, 47], on the final set $+40$ [30, 50] and $+38$ [24, 51].

\begin{table}[!htbp]
  \centering
  \caption{Passes per held-out family (of 9). $^\dagger$: families close to a training family (the same primitives with a different layout or disturbance); the others are new compositions.}
  \label{tab:families}
  \small
  \begin{tabular}{@{}lcccccc@{}}
    \toprule
    & \multicolumn{3}{c}{Evaluator} & \multicolumn{3}{c}{Final set} \\
    \cmidrule(lr){2-4}\cmidrule(lr){5-7}
    Family & H0 & v3, run 1 & v3, run 2 & H0 & v3, run 1 & v3, run 2 \\
    \midrule
    tower & 3 & 7 & 6 & 2 & 5 & 5 \\
    relational & 6 & 6 & 8 & 5 & 7 & 6 \\
    group & 3 & 8 & 7 & 6 & 9 & 8 \\
    line & 2 & 7 & 9 & 2 & 5 & 6 \\
    blocked & 7 & 5 & 7 & 4 & 8 & 8 \\
    reach & 8 & 9 & 9 & 9 & 9 & 9 \\
    clutter$^\dagger$ & 7 & 9 & 9 & 0 & 6 & 6 \\
    crowded$^\dagger$ & 3 & 7 & 6 & 1 & 7 & 8 \\
    kit$^\dagger$ & 3 & 6 & 4 & 3 & 6 & 3 \\
    knocked$^\dagger$ & 0 & 7 & 7 & 1 & 5 & 5 \\
    undone$^\dagger$ & 0 & 8 & 5 & 0 & 6 & 7 \\
    \bottomrule
  \end{tabular}
\end{table}

\section{Human interventions}
\label{app:interventions}

No person wrote into the model, the harness or the procedure during any run. Between runs, we made these changes; each is design work, not self-improvement.
\begin{itemize}\small
  \item Before any improvement run: the request budget (14) and H0 were frozen after a pilot.
  \item After the first version's evaluator result: cost awareness, memory, reruns of infrastructure failures and rounding in the fact gate (second version).
  \item After the second version's evaluator results and the final-set result of its first run: line-by-line fact filtering and three proposals per generation (third version).
  \item After the perception outage in the cross-model test: episodes with a perception connection error count as inconclusive; rechecking all earlier results under this rule changed none of them.
  \item After the first L4 attempt: the robot-agnostic meta-gate and the minimal rewording of $\Phiz$, written down and committed before the second attempt ran.
  \item After the promotion of step 2 but before the second L4 attempt ended and before any run of the held-out check: the design and criteria of the check, committed and not changed afterwards.
\end{itemize}

\section{Unsafe episodes}
\label{app:unsafe}

Across all harnesses and sets we recorded 50 unsafe episodes: 6 on the evaluator, 1 on the final set, 18 on validation, 17 on meta-validation and 5 in training (and 3 in pilot runs outside the five sets). All seven held-out ones came from improved harnesses; their contacts are analysed in \cref{sec:l5}.

\section{Models and settings}
\label{app:settings}

\begin{itemize}\small
  \item \textbf{Executor.} DeepSeek-flash through its API, in the Pi coding-agent kernel with scripted tool calls and the thinking setting ``high''; at most 14 model requests per episode (20 in the budget study).
  \item \textbf{Proposer.} DeepSeek-V4-pro in the runs of the second and third versions in \cref{tab:main}; DeepSeek-flash in the run in which the agent's own model proposes, and as both inner proposer and meta-proposer in \cref{sec:l4}.
  \item \textbf{Perception.} A vision-language model (DeepSeek-flash) names the object in the camera image, with three parallel answers per locate call; depth gives its pose and size.
  \item \textbf{Other executor.} Qwen3.6-35B-A3B, 8-bit, served with llama.cpp on an NVIDIA Jetson AGX Thor.
  \item \textbf{Schedule.} Four generations per run in \cref{sec:l3}; two generations per inner run and three outer steps in \cref{sec:l4}; the validation, meta-validation and held-out sets three repeats per harness, training once per generation.
  \item \textbf{Sets and seeds.} Training, validation and meta-validation take the first four seeds whose scene fits the workcell of each of six families from seeds 0--99, 100--199 and 200--299; the evaluator and the final set the first three such seeds of each of eleven families from 1000--1999 and 2000--2999.
  \item \textbf{Statistics.} Paired by task; 90\% bootstrap intervals over tasks with 2000 resamples.
  \item \textbf{Simulator.} MuJoCo, with the model of the arm pinned to one commit; episodes carry a fingerprint of the simulator source.
  \item \textbf{Scale.} In all, the experiments recorded 12{,}528 episodes on the five sets, 12{,}215 of them conclusive, with 142{,}053 executor model requests, plus the proposer's requests.
  \item \textbf{Availability.} Code and prompts will be released at \coderepo; the texts the paper relies on (both method texts, a learned harness) are given verbatim in \cref{app:method,app:harness}.
\end{itemize}